%% file: main.tex
\documentclass[
]{ceurart}

\usepackage{listings}
\usepackage{graphics} 
\usepackage{epsfig} 
\usepackage{mathptmx} 
\usepackage[utf8]{inputenc}
\usepackage{pifont}     
\usepackage{lipsum}
\usepackage{placeins}
\usepackage{subcaption}
\usepackage{textcomp}
\usepackage{lineno} 
\newcommand{\codelink}{
    \href{https://github.com/ZW471/GraphAU-Pain}{github.com/ZW471/GraphAU-Pain}
}

\hypersetup{
    colorlinks=true,
    linkcolor=MidnightBlue,       
    citecolor=MidnightBlue,      
    urlcolor=gray      
}

\newif\ifcameraready
\camerareadytrue 

\begin{document}


\copyrightyear{2025}
\copyrightclause{Copyright for this paper by its authors.
  Use permitted under Creative Commons License Attribution 4.0
  International (CC BY 4.0).}

\conference{MiGA@IJCAI25: International IJCAI Workshop on 3rd Human Behavior Analysis for Emotion Understanding, August 29, 2025, Guangzhou, China.}

\title{GraphAU-Pain: Graph-based Action Unit Representation for Pain Intensity Estimation}

\ifcameraready
\author[1]{Zhiyu Wang}[%
orcid=0009-0001-8938-3663,
email=zw471@cam.ac.uk,
]
\address[1]{Department of Computer Science, University of Cambridge, Cambridge, UK}

\author[2]{Yang Liu}[%
orcid=0000-0003-2157-0080,
email=yang.liu@oulu.fi,
]
\address[2]{Center for Machine Vision and Signal Analysis, University of Oulu, Oulu, Finland}
\cormark[1]

\author[1]{Hatice Gunes}[%
orcid=0000-0003-2407-3012,
email=hg410@cam.ac.uk
]
\fi

\cortext[1]{Corresponding author. Y. Liu contributed to this work while he was a visiting postdoctoral researcher at the AFAR Lab, Department of Computer Science and Technology, University of Cambridge, UK.}

\begin{abstract}
    Understanding pain-related facial behaviors is essential for digital healthcare in terms of effective monitoring, assisted diagnostics, and treatment planning, particularly for patients unable to communicate verbally. Existing data-driven methods of detecting pain from facial expressions are limited due to interpretability and severity quantification. To this end, we propose GraphAU-Pain, leveraging a graph-based framework to model facial Action Units (AUs) and their interrelationships for pain intensity estimation. AUs are represented as graph nodes, with co-occurrence relationships as edges, enabling a more expressive depiction of pain-related facial behaviors.
    By utilizing a relational graph neural network, our framework offers improved interpretability and significant performance gains. Experiments conducted on the publicly available UNBC dataset demonstrate the effectiveness of the GraphAU-Pain, achieving an F1-score of $66.21\%$ and accuracy of $87.61\%$ in pain intensity estimation. 
    \ifcameraready The code is available for re-implementation at \codelink.\fi
\end{abstract}


\maketitle

\section{Introduction}\label{sec:introduction}
\input{introduction}

\section{Related Work}\label{sec:related-work}
\input{related-work}

\section{Material and Methods}\label{sec:materials-and-methods}
\input{materials-and-methods}

\section{Experiments}\label{sec:experiments}
\input{experiments}

\section{Conclusion}\label{sec:conclusion}
\input{conclusion}

\ifcameraready
\begin{acknowledgments}
  Y. Liu's work was supported in part by the Finnish Cultural Foundation for North Ostrobothnia Regional Fund under Grant 60231712, and in part by the Instrumentarium Foundation under Grant 240016.
\end{acknowledgments}
\fi

\section*{Declaration on Generative AI}
The author(s) have not employed any Generative AI tools.

\bibliography{main}


\end{document}

%% file: introduction.tex
Pain detection is critical in clinical and caregiving settings for timely assessment and improved patient management. Current methods like self-reports and observational evaluations have notable limitations~\cite{Hassan2021AutomaticSurvey}. Self-reports are subjective and depend on patient communication ability, often impaired in nonverbal individuals, children, or those with cognitive impairments. Observational methods, while more objective, require extensive training to ensure accuracy and consistency.

Deep learning has driven interest in automated pain estimation via facial expression analysis, bypassing advanced medical equipment while providing objective measures. Methods like CNNs~\cite{Othman2021AutomaticDatabase, Rodriguez2022DeepClassification} and hybrid frameworks~\cite{ElMorabit2021AutomaticArchitectures, Yang2018IncorporatingEstimation, Barua2022AutomatedImages} have been applied for pain estimation using facial features. Transformers~\cite{Liu2024, Xu2021} perform remarkably in pain prediction from facial videos. However, most rely solely on image features, neglecting physiological insights, limiting clinical interpretability. Additionally, undersampling to address dataset imbalance reduces generalizability in diverse populations~\cite{BenAoun2024ALearning, Chen2018AutomatedReview}.

Pain estimation can leverage features tied to facial expressions. Landmark-based methods like nose tip or eye corner coordinates outperform pixel-based techniques but lack clear physiological links to pain~\cite{Feighelstein2023ExplainableCats, Huang2022HybNet:Estimation}. The Facial Action Coding System (FACS)~\cite{Ekman1978FacialSystem} maps facial movements into Action Units (AUs) with intensity levels, aggregating into the Prkachin and Solomon Pain Intensity (PSPI) score~\cite{Prkachin2008ThePain}. The UNBC dataset~\cite{Lucey2011PainfulDatabase} provides AU and PSPI-labeled videos, supporting AU-informed methods like K-Nearest Neighbor~\cite{Zafar2014PainUnits} and Bayesian Networks~\cite{Guo2021PainNetwork}, which report high accuracy but suffer from overoptimism due to class imbalance and overlook AU relationships. Recent approaches like GLA-CNN~\cite{WU2024100260} and Multiple Instance Learning~\cite{Chen2022LearningLearning} have improved AU relation modeling. However, these methods rely on pixel-wise AU relationships, struggling with subtle facial changes and generalization across diverse racial groups, limiting real-world applicability.

Recent advances in graph neural networks (GNNs), such as Multi-dimensional Edge Feature-based AU Relation Graph for AU (ME-GraphAU)~\cite{Luo2022LearningRecognition} and Graph Relation Network (GRU)~\cite{Wang2020}, have demonstrated promising AU prediction performance on the DISFA dataset~\cite{Mavadati2013DISFA:Database} and BP4D dataset~\cite{Zhang2014BP4D-Spontaneous:Database}. These methods use CNN backbones like ResNet~\cite{He2016} or VGG~\cite{Liu2015} to learn image features for each AU, then construct a relational graph based on the AU features. Each node in such graphs represents an AU, and each edge represents the relationship between a pair of AUs~\cite{liu2022graph}. Through message propagation in GNN layers, the output AU features can capture individual AU information from neighbors as well as structural information~\cite{liu2022uncertain}. Such graph-based AU features can then be aggregated to build a full-face representation informed by AUs and their relationships for downstream tasks~\cite{10214365}, inspiring us to incorporate graph-based AU detection into pain intensity estimation.

Motivated by the above, we introduce \textbf{GraphAU-Pain} for accurate and interpretable pain intensity estimation and summarize our three contributions as follows:
\begin{itemize}
    \item \textbf{GraphAU-Pain Model.} We propose a novel graph-based framework that transforms AU detection into pain estimation by modeling AU relationships as a dynamic graph structure, enabling more expressive and interpretable pain assessment compared to traditional image-driven approaches.
    \item \textbf{Cross-Dataset Transfer with Relabeling.} We introduce a novel transfer learning strategy that leverages DISFA-pretrained weights to address UNBC's limited training data. By creating a hybrid UNBC+ dataset that combines original annotations with predicted labels for missing AUs, we enable effective knowledge transfer while preserving dataset integrity, significantly improving model performance in AU occurrence prediction.
    \item \textbf{High Performance \& Interpretability.} Comprehensive experiments validate GraphAU-Pain for pain estimation, outperforming GLA-CNN~\cite{WU2024100260} (current state-of-the-art work in this area) and demonstrating improved interpretability via explicit AU modeling.
\end{itemize}

%% file: related-work.tex
\subsection{Facial Action Unit Detection}\label{subsec:facial-au-detection}

FACS~\cite{Ekman1978FacialSystem} categorizes facial expressions into 44 fundamental components known as AUs, each corresponding to specific muscle movements with intensity ratings from 0 to 5. For instance, AU1 represents the 
``Inner Brow Raiser'' and AU2 the ``Outer Brow Raiser.'' These AUs provide a systematic method for analyzing facial expressions, including pain and other affective states. Early approaches to AU detection relied on traditional machine learning methods, with OpenFace~\cite{Baltrusaitis2018OpenFaceToolkit} being a widely used open-source tool based on SVMs~\cite{Baltrusaitis2015Cross-datasetDetection}. Deep learning methods like Deep Region and Multi-label Learning~\cite{Zhao2016DeepDetection} established early benchmarks but failed to model AU interdependencies. Later attention-based methods~\cite{Shao2022FacialLearning, MiriamJacob2021FacialTransformers} improved performance significantly, though their pixel-wise approach faced limitations in capturing subtle facial changes and generalizing across diverse racial groups.

Graph Neural Networks (GNNs) have emerged as a powerful architecture for modeling complex dependencies between facial landmarks or AUs. A GNN layer passes information between nodes through edges, allowing nodes to learn their neighborhood information. This makes GNNs essential for learning relational representations, as demonstrated in fields like knowledge graph construction~\cite{zuoKG4DiagnosisHierarchicalMultiagent2025} and recommender systems~\cite{Wu2022GraphSurvey}. In facial AU analysis, some studies introduced prior AU co-occurrence knowledge via Graph Convolutional Networks~\cite{He2021FacialNetworks, Jia2023ARecognition}, while others like Graph Relation Network (GRN)~\cite{Wang2020} explored knowledge-free approaches. GRN constructs a fully connected directed graph with image features as nodes and uses attention-based edge functions, achieving MAE of $0.7$ on BP4D and $0.2$ on DISFA. 
Unlike GRN, which fully connects all nodes, ME-GraphAU~\cite{Luo2022LearningRecognition} introduces a novel approach by using a CNN backbone to extract features for each node that represents an AU and establishing connections between nodes based on feature similarity. This architecture effectively captures facial feature relationships by modeling AUs as nodes and their interactions as edges, achieving strong performance with F1 scores of $65.5\%$ on BP4D and $63.1\%$ on DISFA.
Despite these advances, direct adaptation of AU detection models to pain estimation tasks remains challenging due to two key limitations. First, pain estimation datasets are typically much smaller than those used for AU detection. Second, these datasets exhibit a significant class imbalance in AU occurrences. Our GraphAU-Pain framework addresses both challenges through a novel transfer learning strategy, as detailed in Sec.~\ref{subsec:pretraining-and-fine-tuning}.

\subsection{Pain Prediction Based on Facial Expressions}\label{subsec:pain-detection}

Early approaches relied on feature extraction and classification techniques like KNN~\cite{Zafar2014PainUnits} and Random Forest~\cite{XingZhang2015ThreeAnalysis}. These handcrafted feature approaches were limited by variations in head pose, lighting, and spontaneous expressions.

Recent deep learning approaches have evolved from traditional CNNs~\cite{Wang2017RegularizingRegression, Zhou2016RecurrentVideo} to more sophisticated architectures that incorporate AU features. While CNNs provide a foundation through pixel-level analysis, their lack of physiological knowledge limits both interpretability and generalizability. This limitation has driven the development of more advanced approaches, such as LSTM-based continuous pain monitoring~\cite{Othman2023ClassificationDatabase}. However, this promising work faces practical constraints due to its reliance on a private dataset~\cite{Gruss2019Multi-ModalStimuli}. Similarly, while AU-based pain prediction has shown potential on BP4D~\cite{Feghoul2023FacialDetection}, its binary classification approach may not capture the nuanced spectrum of natural pain experiences.

The state-of-the-art GLA-CNN~\cite{WU2024100260} represents a significant advancement by combining CNNs with attention mechanisms to analyze facial pain and AU relationships. On the UNBC dataset, it achieves $36.2\%$ F1-score and $56.5\%$ accuracy. While these metrics show improvement, they remain insufficient for clinical applications. The model's fine-grained category scheme for pain levels fails to account for PSPI's sensitivity to AU intensity. This limitation becomes apparent when considering cases such as intense AU4 (brow lowering) can occur without pain~\cite{Werner2017AutomaticDescriptors}, leading to confusion between its \texttt{Weak Pain} ($PSPI=0$), \texttt{Weak Pain} ($PSPI=1$), and \texttt{Mild Pain} ($PSPI=2$) categories.
Furthermore, the black-box nature of the design limits clinical utility by obscuring how the estimated pain intensity is derived.
These challenges highlight the need for more reliable and interpretable pain detection models, motivating our development of GraphAU-Pain. Our approach explicitly incorporates AUs through a graph-based modeling framework, providing enhanced interpretability through transparent AU-pain intensity relationships and improved accuracy via comprehensive modeling of AU interdependencies.

%% file: materials-and-methods.tex
\subsection{Implementation Overview}\label{subsec:implementation-overview}
The GraphAU-Pain model is designed to estimate pain intensity based on facial image data, and its training involves two key steps. First, the full-face and AU representation learning modules are trained for \emph{AU occurrence prediction}.
These modules are adapted from the AU Relationship-aware Node Feature Learning (ANFL) component of ME-GraphAU~\cite{Luo2022LearningRecognition}.
Because directly training ANFL for AU occurrence prediction on the UNBC dataset only yielded a $20\%$ average F1-score, we developed a cross-dataset transfer learning strategy to improve performance. This strategy transfers AU prediction capabilities from the DISFA dataset model trained by Luo et al.~\cite{Luo2022LearningRecognition} to the UNBC dataset. To address differences in AU labels between datasets, we created a relabeled UNBC+ dataset to align the AU annotations and used undersampling to address AU imbalance.
Second, with the weights trained for AU prediction, GraphAU-Pain is then trained for \emph{pain intensity estimation} with the full UNBC dataset. Facial images are first processed through the CNN backbone to extract pixel-wise full-face representations. These representations are then transformed into graph-based structures to learn AU-specific features. Finally, the full-face and AU-based features are combined to estimate pain intensity. The performance of the model is evaluated using two metrics: average F1-score and accuracy.

\begin{figure}[!t]
    \centering
    \includegraphics[width=0.85\columnwidth]{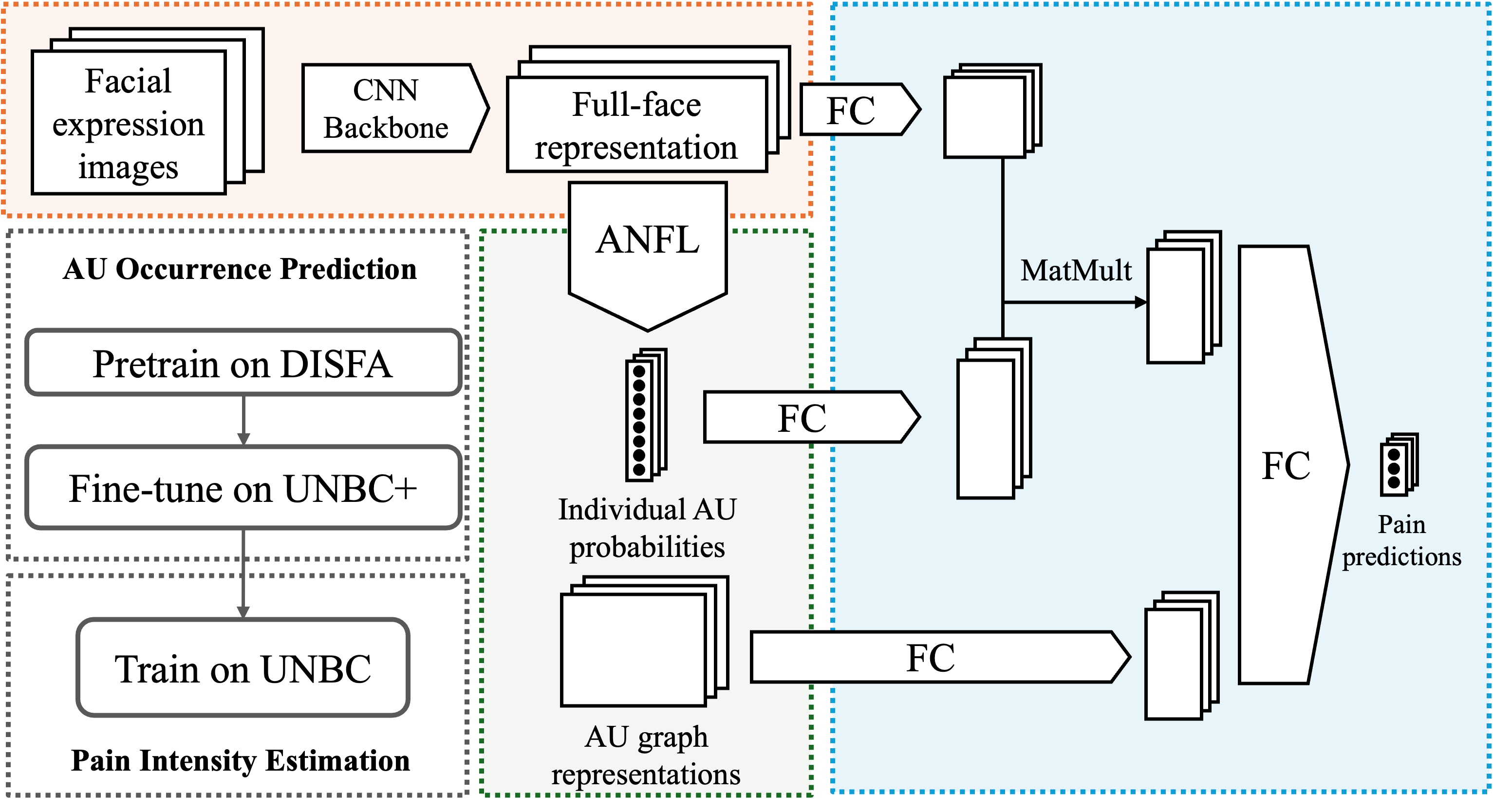}
    \caption{Illustration of the GraphAU-Pain model architecture. A global CNN backbone (orange) extracts a full-face representation, which is split into three feature branches by the AU representation learning module (green) and processed by the pain level classification module (blue). \textit{ANFL} denotes the AU Relationship-aware Node Feature Learning component, \textit{FC} the fully-connected layer, and \textit{MatMult} matrix multiplication. GraphAU-Pain is trained by first initializing weights of the representation learning modules with the AU occurrence prediction task and then train the full model for the pain intensity estimation task.}
    \label{fig:model-architecture}
\end{figure}

As illustrated in Fig.~\ref{fig:model-architecture}, the GraphAU-Pain model comprises three sequential modules: \emph{Full-face Representation Learning} that extracts high-level facial features through pixel-based global representations, \emph{AU Representation Learning} that captures both local and global AU using graph-based representations, and a \emph{Pain Intensity Classifier} that maps the learned features to specific pain intensity levels.

\subsection{Model}\label{subsec:model-architechture}

\paragraph{Full-face Representation Learning}
GraphAU-Pain uses a ResNet-50 backbone for extracting a full-face representation from an input image. By inputting a face image \(\mathbf{x} \in \mathbb{R}^{172 \times 172 \times 3}\) to the backbone, we obtain \(\mathbf{h_{b}}\in \mathbb{R}^{36 \times 2048}\), which represents 36 facial image features of length 2048, each corresponding to a position in the image.

\paragraph{AU Representation Learning}
After acquiring the backbone feature \(\mathbf{h_{b}}\), the AU representation module learns two distinct embeddings: \(\mathbf{h_{a}}\in \mathbb{R}^{d_{\text{AU}}}\) (\(d_{\text{AU}} = 512\)) encoding individual AU occurrences, and \(\mathbf{h_{g}}\in \mathbb{R}^{d_{\text{AU}}}\) encoding the complete AU relational graph structure. The module first transforms \(\mathbf{h_{b}}\) through \(n_{\text{AU}}\) fully connected layers to generate initial AU representations $\mathbf{H_a} \in \mathbb{R}^{n_{\text{AU}} \times d_{\text{AU}}}$, where each row represents one AU. These representations serve as node features in a graph where each node connects to its \(K=3\) most similar nodes based on dot-product similarity. The graph structure is then processed through a graph convolutional layer:
\begin{align}
        \mathbf{H_a'} \;=\; \text{ReLU} \Bigl( \mathbf{H_a} + \text{BN}\bigl(\mathbf{A}^{\top}\text{FC}_{1}(\mathbf{H_a}) + \text{FC}_{2}(\mathbf{H_a})\bigr) \Bigr)  \;\in\; \mathbb{R}^{n_{\text{AU}} \times d_{\text{AU}}},
\end{align}
where BN denotes batch normalization, FC represents fully connected layers, and $\mathbf{A}$ is the normalized adjacency matrix. The graph representation $\mathbf{h_{g}} \in \mathbb{R}^{d_{\text{AU}}}$ is obtained through global sum pooling of the node embeddings. Although we could also add an edge update module here~\cite{Luo2022LearningRecognition}, we omitted it to avoid overfitting on the UNBC dataset. For each AU, its occurrence probability $p_i$ is computed as the cosine similarity between its representation $h_{a, i}$ and a learnable vector $s_i$:
\begin{align}
    p_i = \frac{\text{ReLU}(h_{a, i})^{\top}\text{ReLU}(s_{i})}{\|\text{ReLU}(h_{a, i})\|_2 \|\text{ReLU}(s_{i})\|_2},
\end{align}
where $\| \cdot \|$ represents the L2 norm. This AU occurrence prediction can serve as either a pretraining task or an auxiliary training objective alongside the primary pain intensity estimation task. In this work, we use it as a pretraining task on an undersampled dataset to enforce the AU representation module to focus on minority classes.
After obtaining these three features, three fully connected layers with ReLU activation map each of them to a common dimension of $36$, producing ${\mathbf{h_{b}'}}$, ${\mathbf{h_{a}'}}$, and ${\mathbf{h_{g}'}}$. (For \(\mathbf{h_{b}}\), FC is applied row-wise.) Finally, a feature-infusing step on \(\mathbf{h_{a}}'\) and \(\mathbf{h_{b}}'\) is performed:
\begin{align}
    \mathbf{h_{ab}} \;=\; \text{ReLU}(\mathbf{h_{a}'^\top} \,\mathbf{h_{b}}') \;\in\; \mathbb{R}^{36}.
\end{align}

\paragraph{Representation Classifier}
The final pain intensity classification is get by concatenating the interaction scalar \(\mathbf{h_{ab}}\) with the \(\mathbf{h_{g}}'\) feature, then passing the result to an FC layer:
\begin{align}
    \mathbf{\hat{y}} \;=\; \mathbf{W}_{y}\,[\mathbf{h_{ab}}\parallel \mathbf{h_{g}'}] + \mathbf{b}_{y}
    \;\in\; \mathbb{R}^{d_{\text{pain}}},
\end{align}
where \(d_{\text{pain}}=3\) for the one-hot encoding of the three-level pain intensity classification used in this work.

\subsection{Loss Function}\label{subsec:loss-function}
With over 80\% of the frames demonstrating no expression of pain, the class imbalance in the UNBC dataset poses a significant challenge for deep learning models. To minimize this, we employ a weighted cross-entropy loss to prioritize underrepresented classes. It is calculated by
\begin{align}
    \mathcal{L}
    = -\frac{1}{N} \sum_{i=1}^N \sum_{j=1}^C w_j \, y_{i,j} \, \log \max\bigl(p_{i,j}, \epsilon\bigr),
\end{align}
where $N$ is the number of samples, $C$ is the number of classes, \(p_{i,j}\) are the softmax output probabilities for the \(i\)-th sample belonging to the \(j\)-th class, \(y_{i,j}\) is a binary indicator (1 if the \(i\)-th sample belongs to the \(j\)-th class, otherwise 0), \(w_j\) are class weights, and \(\epsilon=1e^{-8}\) prevents \(\log(0)\). The weight of a class $c_j$ is calculated by
\begin{align}
w_{j} = C \cdot \frac{1 / \text{occurrence\_rate}(c_j)}{\sum_{k=1}^{C} 1 / \text{occurrence\_rate}(c_k)}.
\end{align}
In this work, the class weights are $0.07$ for \texttt{No Pain}, $0.33$ for \texttt{Mild Pain}, and $2.6$ for \texttt{Obvious Pain}.

\subsection{Transfer Learning}\label{subsec:pretraining-and-fine-tuning}
Our preliminary experimental results showed that directly training GraphAU-Pain on UNBC yielded unsatisfactory results, with 30--40\% F1-score and 60--80\% accuracy after trying several settings. To tackle this problem and improve AU prediction performance through transfer learning, in this work we initialized the weights of the full-face and AU representation modules with the weights pretrained on DISFA provided by Luo et al.~\cite{Luo2022LearningRecognition}.
 We chose the DISFA dataset~\cite{Mavadati2013DISFA:Database} for pretraining because it is three times larger than UNBC and provides high-quality AU annotations that align well with UNBC's AU annotation scheme---sharing six out of eight AUs with UNBC and three of them are used in PSPI calculation---making this dataset suitable for transfer learning. However, since DISFA includes two additional AUs (AU1 and AU2) not present in UNBC's original annotations, for fine-tuning the pretrained weights, we need to label these additional AUs for UNBC to ensure complete AU coverage. To do this, we pass UNBC's facial images through the pretrained representation learning modules to predict all eight AU labels. We then create a hybrid dataset (UNBC+) by keeping UNBC's original annotations for the six overlapping AUs while using the predicted values for AU1 and AU2. This relabeling process ensures that our model can learn from a complete set of AU activations while maintaining the reliability of UNBC's original annotations where available.

%% file: experiments.tex
\subsection{Datasets and Labels}\label{subsec:datasets-and-labels}
GraphAU-Pain was trained on the UNBC-McMaster Shoulder Pain Expression Archive Database~\cite{Lucey2011PainfulDatabase}. The dataset contains 48,398 colored frames from 25 participants with shoulder problems, showing facial expressions during pain-inducing actions. The faces were detected with OpenCV's \texttt{haarcascade frontalface default} classifier and cropped to $172\times172$. Each frame has 10 AU intensities ($0-5$) and PSPI scores ($0-16$), calculated as \(PSPI = AU4 + \max(AU6, AU7) + \max(AU9, AU10) + AU43\)~\cite{Prkachin2008ThePain}.
PSPI pain intensity is categorized into ordinal levels: \texttt{No Pain} ($PSPI=0$), \texttt{Mild Pain} ($PSPI \in [1,4]$), and \texttt{Obvious Pain} ($PSPI \geq 5$). The categories are distributed in a skewed way, with each category respectively consisting approximately $82\%$, $15\%$, and $3\%$ of the UNBC dataset. This categorization is more clinically meaningful and interpretable compared to the method used by Wu et al.~\cite{WU2024100260}, as it mitigates the high sensitivity of PSPI scores to AU variations. Moreover, we replaced AU intensity with occurrence by capping the AU score at one.

\subsection{Training and Evaluation Details}\label{subsec:training}
To prepare AU occurrence prediction before pain intensity estimation training, the supervised fine-tuning (SFT) process of GraphAU-Pain's AU representation learning module was performed on the ANFL component with weights pretrained on DISFA for 20 epochs provided by Luo et al.~\cite{Luo2022LearningRecognition}.
Our relabeled UNBC+ dataset was used in the SFT process and undersampled to address data imbalance. The undersampling process involved randomly removing approximately $90\%$ of facial images with $PSPI=0$ and excluding facial images without active AUs.
The full list of frames included in this subset is made available in the code repository. We use all \(n_{\text{AU}}=8\) AU labels in the UNBC+ dataset as listed in Table~\ref{tab:au_metrics_single_row}. We trained the module through SFT for 17 epochs with a learning rate of \(1e^{-5}\), a batch size of 16, and an Adam optimizer with \(\beta_1=0.9\), \(\beta_2=0.999\), and a weight decay of \(5e^{-4}\). The AU representation learning module achieved remarkable results in AU detection, compared to state-of-the-art results, as detailed in Table~\ref{tab:au_metrics_single_row}. The training for pain intensity estimation used the same hyperparameters as the SFT process but was performed on the full original UNBC dataset.

\begin{table}[t]
  \centering
  \caption{F1-scores and accuracies achieved for each AU by the fine-tuned representation learning module. The average F1-score and accuracy are calculated by averaging the F1-scores and accuracies of all AUs. While accuracy remains high across all AUs, F1-scores are notably lower for AU9 and AU26, likely due to class imbalance.}
  \label{tab:au_metrics_single_row}
  \resizebox{0.7\columnwidth}{!}{%
    \begin{tabular}{lccccccccc}
      \toprule
      Metric & AU1 & AU2 & AU4 & AU6 & AU9 & AU12 & AU25 & AU26 & Avg \\
      \midrule
      F1 & 63.09 & 88.65 & 70.74 & 88.46 & 31.94 & 92.12 & 63.93 & 6.63  & 63.20 \\
      Acc. & 90.38 & 79.88 & 92.04 & 79.62 & 91.43 & 86.09 & 83.11 & 85.21 & 85.98 \\
      \bottomrule
    \end{tabular}%
  }

\end{table}

The GraphAU-Pain model was trained using a learning rate of \(1e^{-4}\), a batch size of 64, and an Adam optimizer with the same hyperparameters used in SFT. The representation learning module is set to connect edges between an AU node and its 3 most similar AU nodes. The weights learned through SFT on ANFL were used to initialize the ResNet backbone and the representation learning module of GraphAU-Pain. The model was trained on the full UNBC dataset for 8 epochs on an NVIDIA GeForce RTX 4070 GPU (8 GB) and an Intel i7-13900H CPU, with an estimated training time of about 3 minutes per epoch. The evaluation uses the same metrics as the state-of-the-art method GLA-CNN~\cite{WU2024100260}: accuracy, average F1, average recall, and average precision, where all average values are unweighted.

\begin{table}[!t]
  \centering
  \caption{GraphAU-Pain's performance statistics. Our model achieves strong performance for \texttt{No Pain} detection but lower and similar performance for \texttt{Mild} and \texttt{Obvious} pain categories, likely due to data imbalance. The overall F1/Recall/Precision/Accuracy are calculated by averaging the F1/Recall/Precision/Accuracy of all categories.}
  \label{tab:classification_results_cat3}
  \resizebox{0.45\columnwidth}{!}{%
    \begin{tabular}{lcccc}
      \toprule
      Metric & No Pain & Mild & Obvious & Overall \\
      \midrule
      F1-score & 93.10 & 51.19 & 54.35 & 66.21 \\
      Recall & 92.85 & 52.48 & 51.02 & 65.45 \\
      Precision & 93.35 & 49.96 & 58.14 & 67.15 \\
      Accuracy & - & - & - & 87.61 \\
      \bottomrule
    \end{tabular}%
  }
\end{table}

\subsection{Model Performance}\label{subsec:model-performance}
\begin{figure}[!t]
  \centering
  \begin{subfigure}[b]{0.38\columnwidth}
    \includegraphics[width=\columnwidth]{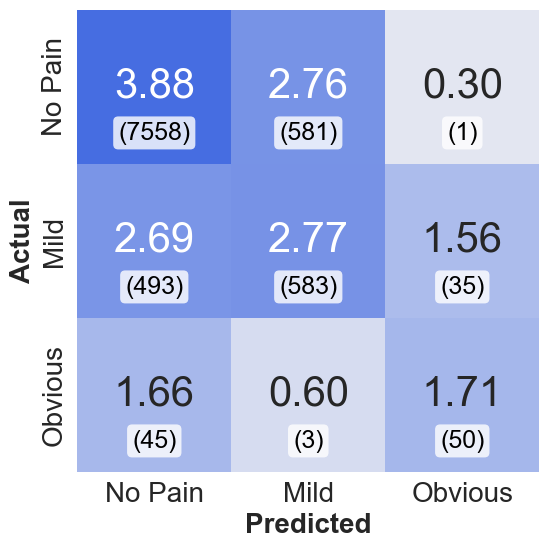}
    \caption{Classifying 3 Pain Intensities}
    \label{fig:confusion-matrix-full-graph-sub1}
  \end{subfigure}
  \begin{subfigure}[b]{0.38\columnwidth}
    \includegraphics[width=\columnwidth]{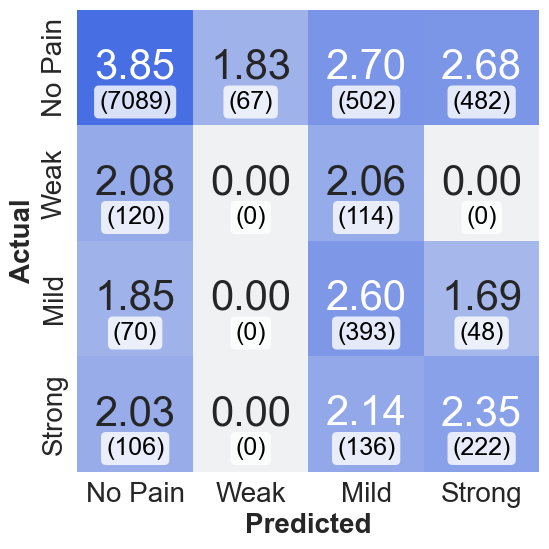}
    \caption{Classifying 4 Pain Intensities}
    \label{fig:confusion-matrix-full-graph-sub2}
  \end{subfigure}
  \caption{Confusion matrix of GraphAU-Pain. The numbers are transformed in $\log_{10}$ scale for better visualization; the original number is shown below the scaled one. The 3-category classification (left) is introduced in this work to mitigate the excessive sensitivity of the 4-category system (right) proposed by Wu et al.~\cite{WU2024100260}. The 4-category system's additional ``Weak Pain'' category (PSPI=1) proves too sensitive to AU intensity variations and lacks clinical significance when separated from our ``Mild Pain'' category. Therefore, the lower performance in recognizing ``Weak Pain'' reflects the metric's sensitivity to minor AU variations rather than model limitations.}
  \label{fig:confusion-matrix-full-graph}
\end{figure}

Overall, GraphAU-Pain achieves a commendable average F1-score of $66.21\%$ and a high accuracy of $87.61\%$. Table~\ref{tab:classification_results_cat3} details the per-class results, showing strong performance for the \texttt{No Pain} category, which has an F1-score of $93.10\%$. However, performance declines for the \texttt{Mild} and \texttt{Obvious} categories, reflecting F1-scores of $51.19\%$ and $54.35\%$, respectively. This performance gap between \texttt{No Pain} and the \texttt{Mild/Obvious} categories is largely attributable to the dataset's pronounced class imbalance. The log-scaled confusion matrix in Figure~\ref{fig:confusion-matrix-full-graph} further illustrates the distribution of predictions. While \texttt{No Pain} dominates the diagonal, indicating high accuracy there, some off-diagonal misclassifications occur between \texttt{Mild} and \texttt{Obvious}, and there is a noticeable bias toward predicting \texttt{No Pain}. Consequently, while the model is robust in detecting \texttt{No Pain}, additional strategies are needed to better distinguish between higher pain intensities.

\subsection{Comparison to SOTA}\label{subsec:model-comparision}
To the best of our knowledge, GLA-CNN~\cite{WU2024100260} is the only other method that uses AUs for pain-intensity estimation on UNBC while focusing on cross-sectional facial image frames. No additional methods apply exactly the same categorization scheme, so we compare GraphAU-Pain with GLA-CNN and other models reported in~\cite{WU2024100260}. To align labels, we reclassify pain intensities into four categories: \texttt{No Pain} ($PSPI=0$), \texttt{Weak Pain} ($PSPI=1$), \texttt{Mild Pain} ($PSPI=2$), and \texttt{Strong Pain} ($PSPI\geq3$). By altering only the model's final layer from three to four outputs and keeping other settings unchanged, GraphAU-Pain shows substantial gains in both accuracy and average F1-score, as shown in Table~\ref{tab:model_comparison} and Figure~\ref{fig:confusion-matrix-full-graph-sub2}. Note that GLA-CNN and the other compared models were trained and evaluated on an undersampled subset of UNBC~\cite{WU2024100260} to deal with class imbalance, whereas GraphAU-Pain is trained on the full dataset. Therefore, their published performance might be higher than what would have been achieved on the full dataset.
\begin{table}[!t]
  \centering
  \caption{Comparison of GraphAU-Pain with other models on the UNBC dataset. We compare our results with those reported by Wu et al.~\cite{WU2024100260}, and we retrained our model with the 4-category system for fair comparison.}
  \label{tab:model_comparison}
  \resizebox{0.65\columnwidth}{!}{%
    \begin{tabular}{lcccc}
      \toprule
      Model & Accuracy & F1-score & Recall & Precision \\
      \midrule
      GraphAU-Pain (Ours) & 82.40 & 43.83 & 52.96 & 39.96 \\
      GLA-CNN~\cite{WU2024100260} & 56.45 & 36.52 & 34.08 & 43.23 \\
      MA-NET~\cite{9474949} & 55.16 & - & - & - \\
      Swin Transformer~\cite{9878941} & 54.40 & - & - & - \\
      Attention CNN~\cite{10.1371/journal.pone.0232412} & 51.10 & - & - & - \\
      LBP~\cite{7820930} & 34.70 & - & - & - \\
      Head Analysis~\cite{6617456} & 28.10 & - & - & - \\
      LPQ~\cite{7820930} & 23.40 & - & - & - \\
      \bottomrule
    \end{tabular}%
  }
\end{table}

\subsection{Ablation Analysis}\label{subsec:ablation-analysis}
The ablation analysis in Table~\ref{tab:all_models_results} underscores the critical role of graph representation and GNN in the GraphAU-Pain model. Removing the graph representation (\textit{w/o graph rep.}) reduces the mean F1-score from $66.2\%$ to $63.1\%$, mainly due to the performance drops in \texttt{No Pain} and \texttt{Mild Pain}, highlighting the value of graph modeling for capturing AU relationships. Similarly, removing the GNN layer (\textit{w/o GNN}) causes a significant drop to $40.3\%$, emphasizing the importance of graph-based interactions in learning AU features. The simplest setup (\textit{Only ResNet}), relying solely on CNNs, achieves the lowest mean F1-score of $35.2\%$, demonstrating that CNNs alone fail to effectively model AU correlations for pain estimation. These results affirm the superiority of graph-based learning methods for pain estimation.

\begin{table}[t]
  \centering
  \caption{Ablation analysis for GraphAU-Pain. \textit{w/o graph rep.} removes the graph representation from the model. \textit{w/o GNN} removes the GNN layer. \textit{Only ResNet} relies solely on CNNs. The highest value for each category and metric is \textbf{bold}; the second highest is \underline{underlined}.}
  \resizebox{\columnwidth}{!}{%
    \begin{tabular}{l*{4}{c}*{4}{c}*{4}{c}}
      \toprule
      & \multicolumn{4}{c}{F1}
      & \multicolumn{4}{c}{Precision}
      & \multicolumn{4}{c}{Recall} \\
      \cmidrule(lr){2-5}\cmidrule(lr){6-9}\cmidrule(lr){10-13}
      Model
      & No Pain & Mild & Obvious & Mean
      & No Pain & Mild & Obvious & Mean
      & No Pain & Mild & Obvious & Mean \\
      \midrule
      Full
      & \textbf{93.1} & \textbf{51.2} & \underline{54.3} & \textbf{66.2}
      & 93.4          & \textbf{50.0} & \underline{58.1} & \textbf{67.2}
      & \textbf{92.9} & 52.5          & \textbf{51.0} & \underline{65.4} \\
      w/o graph rep.
      & \underline{87.7} & \underline{46.4} & \textbf{55.2} & \underline{63.1}
      & \textbf{95.9}    & \underline{33.9} & \textbf{60.2}    & \underline{63.4}
      & \underline{80.8} & \textbf{73.4}   & \textbf{51.0}   & \textbf{68.4}   \\
      w/o GNN
      & 86.4 & 30.7 & 4.0  & 40.3
      & \underline{93.9} & 23.8 & 2.5  & 40.1
      & 79.9  & 43.2 & 10.2 & 44.4           \\
      Only ResNet
      & 70.3 & 31.0 & 4.2  & 35.2
      & 92.8  & 20.1 & 2.5  & 38.5
      & 56.5  & \underline{68.3} & 15.3 & 46.7   \\
      \bottomrule
    \end{tabular}%
  }
  \label{tab:all_models_results}
\end{table}

\subsection{Discussion and Future Work}\label{subsec:discussions}

GraphAU-Pain demonstrates significant potential for advancing automated pain assessment through several key contributions. By leveraging graph-based learning to model AU relationships, our approach achieves superior performance compared to existing methods, with an accuracy of 87.61\% in the clinically meaningful three-category classification system. The model's strong performance in detecting \texttt{No Pain} (93.10\% F1-score) makes it particularly valuable for initial screening applications. Furthermore, the AU representation learning module provides a more interpretable framework for understanding how different facial expressions contribute to pain assessment. This could lead to more reliable and explainable automated pain monitoring systems in clinical settings, potentially reducing the burden on healthcare providers and improving patient care through continuous, objective pain assessment.

While the proposed method shows promising results in pain estimation, there remains room for improvement.
Firstly, since PSPI only captures facial expressions, it may not reflect true subjective pain~\cite{DeSario2023UsingReview}. Future research could focus on finding alternative pain indicators. Secondly, aligning UNBC with DISFA through UNBC+ removes three pain-related AUs and adds noise through predicted labels, potentially impacting performance. A promising direction is to design AU occurrence prediction models specifically for pain-oriented datasets like UNBC. Lastly, while the AU occurrence-based representation learning module provides satisfactory representation, AU intensity-based approaches (e.g., GRN~\cite{Wang2020}) could also be explored since AU intensity better relates to pain intensity. However, this direction may require more complex models and additional training data to mitigate the impact of data imbalance.

%% file: conclusion.tex
This paper presents GraphAU-Pain, a GNN-based model combining graph-based AU features with full-face representation for pain prediction. It surpasses the state-of-the-art methods while enabling AU-informed pain estimation for clinical transparency. GraphAU-Pain addresses challenges like limited data and class imbalance in the UNBC dataset through a novel transfer learning strategy. Key contributions include improved pain classification benchmarks, better interpretability through AU-based representations, and critical baselines for future AU-based pain estimation. Overall, this work demonstrates the potential of GNNs for accurate, clinically viable pain estimation solutions.